\documentclass[table]{article}

\usepackage[table, dvipsnames]{xcolor}
\usepackage[font=small, labelfont=bf]{caption}
\usepackage{multicol}
\usepackage[bookmarks=true, pagebackref]{hyperref}
\usepackage[square, numbers, sort&compress]{natbib}
\usepackage{lipsum}
\usepackage{xspace}
\usepackage{todonotes}
\usepackage{amsmath}
\usepackage{amssymb}
\usepackage{booktabs}
\usepackage{algorithmic}
\usepackage[toc,page,header]{appendix}
\usepackage{minitoc}
\usepackage[ruled, vlined, linesnumbered]{algorithm2e}
\usepackage{wrapfig}
\usepackage{array}
\usepackage{soul}
\usepackage{subcaption}
\usepackage{multirow}

\def\eqref#1{equation~\ref{#1}}

\def\1{\bm{1}}

\DeclareMathAlphabet{\mathsfit}{\encodingdefault}{\sfdefault}{m}{sl}
\SetMathAlphabet{\mathsfit}{bold}{\encodingdefault}{\sfdefault}{bx}{n}

\def\gG{{\mathcal{G}}}

\usepackage[preprint]{corl_2022} %

\title{LM-Nav: Robotic Navigation with Large Pre-Trained \\ Models of Language, Vision, and Action}

\newcommand{\sysName}[0]{LM-Nav\xspace}

\author{
  Dhruv Shah$^{\dagger\beta}$, Błażej Osiński$^{\dagger\beta\omega}$, Brian Ichter$^\gamma$, Sergey Levine$^{\beta\gamma}$ \\ 
  $^\beta$UC Berkeley, $^\omega$University of Warsaw, $^\gamma$Robotics at Google
}

\newcommand{\llm}[0]{\textbf{LLM}\xspace}
\newcommand{\vlm}[0]{\textbf{VLM}\xspace}
\newcommand{\gcc}[0]{\textbf{VNM}\xspace}
\newcommand{\gcp}[0]{\textbf{VNM}\xspace}

\newcommand{\projectwebsite}{\href{https://sites.google.com/view/lmnav}{\texttt{sites.google.com/view/lmnav}}}

\begin{document}
\maketitle
\doparttoc %
\faketableofcontents %

\newcommand\blfootnote[1]{%
  \begingroup
  \renewcommand\thefootnote{}\footnote{#1}%
  \addtocounter{footnote}{-1}%
  \endgroup
}

\begin{abstract}
Goal-conditioned policies for robotic navigation can be trained on large, unannotated datasets, providing for good generalization to real-world settings. However, particularly in vision-based settings where specifying goals requires an image, this makes for an unnatural interface. Language provides a more convenient modality for communication with robots, but contemporary methods typically require expensive supervision, in the form of trajectories annotated with language descriptions. We present a system, LM-Nav, for robotic navigation that enjoys the benefits of training on unannotated large datasets of trajectories, while still providing a high-level interface to the user. Instead of utilizing a labeled instruction following dataset, we show that such a system can be constructed entirely out of pre-trained models for navigation (ViNG), image-language association (CLIP), and language modeling (GPT-3), without requiring any fine-tuning or language-annotated robot data. 
We instantiate LM-Nav on a real-world  mobile robot and demonstrate long-horizon navigation through complex, outdoor environments from natural language instructions.%
\end{abstract}


\blfootnote{$^\dagger$ These authors contributed equally, order decided by a coin flip. Check out the project page for experiment videos, code, and a user-friendly Colab notebook that runs in your browser: \projectwebsite
}

\section{Introduction}
\label{sec:intro}

One of the central challenges in robotic learning is to enable robots to perform a wide variety of tasks on command, following high-level instructions from humans. This requires robots that can understand human instructions, and are equipped with a large repertoire of diverse behaviors to execute such instructions in the real world. Prior work on instruction following in navigation has largely focused on learning from trajectories annotated with textual instructions~\cite{anderson2018vision, gu2022vision, ku2020rxr, jain2019stay, yan2019xl}. This enables understanding of textual instructions, but the cost of data annotation impedes wide adoption. On the other hand, recent work has shown that learning robust navigation is possible through goal-conditioned policies trained with self-supervision. These utilize large, unlabeled datasets to train vision-based controllers via hindsight relabeling~\cite{Manderson2010SelfSupervisedGP, boris2006improving, gandhi2017fly, kouris2018fly, kahn2018gcg, shah2020ving}. They provide scalability, generalizability, and robustness, but usually involve a clunky mechanism for goal specification, using locations or images. In this work, we aim to combine the strengths of both approaches, enabling a self-supervised system for robotic navigation to execute natural language instructions by leveraging the capabilities of pre-trained models \emph{without any user-annotated navigational data}. Our method uses these models to construct an ``interface'' that humans can use to communicate desired tasks to robots. This system enjoys the impressive generalization capabilities of the pre-trained language and vision-language models, enabling the robotic system to accept complex high-level instructions.

\begin{figure}[t]
    \centering
    \includegraphics[width=\linewidth]{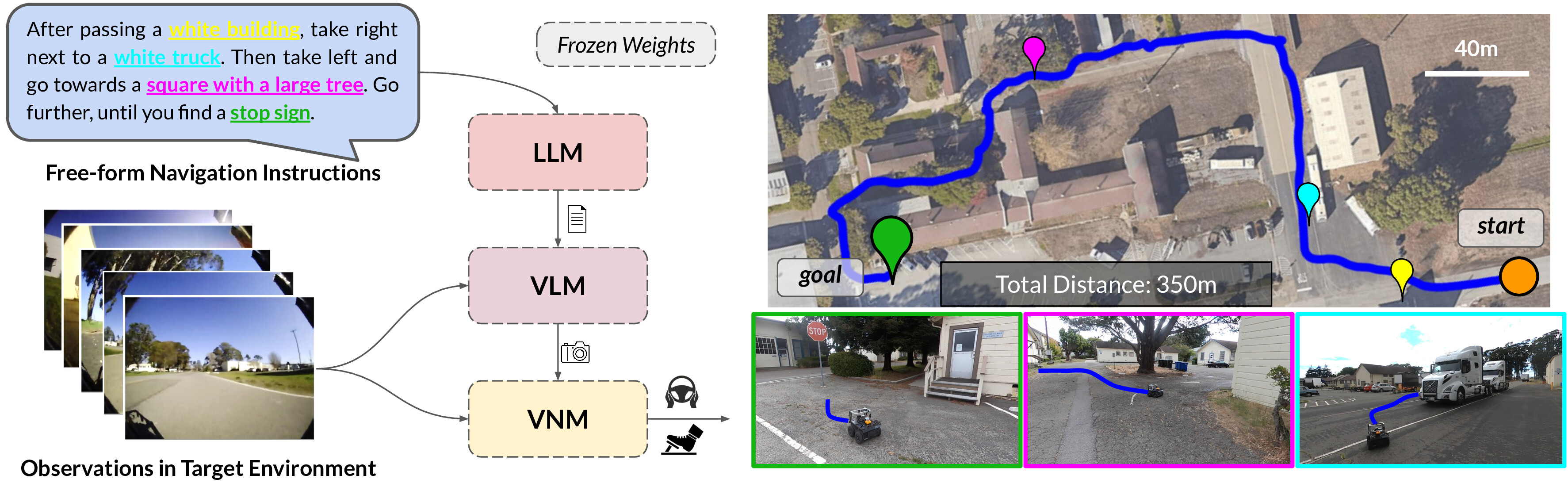}
    \caption{\textbf{Embodied instruction following with \sysName:} Our system takes as input a set of raw observations from the target environment and free-form textual instructions (left), deriving an actionable plan using three \emph{pre-trained} models: a large language model (\llm) for extracting landmarks,
    a vision-and-language model (\vlm) for grounding, and a visual navigation model (\gcc) for execution. This enables \sysName to follow textual instructions in complex environments purely from visual observations (right) \emph{without any fine-tuning}.}
    \label{fig:overview}
\end{figure}

Our main observation is that we can utilize off-the-shelf \emph{pre-trained models} trained on large corpora of visual and language datasets --- that are widely available and show great few-shot generalization capabilities --- to create this interface for embodied instruction following.  To achieve this, we combine the strengths of two such robot-agnostic pre-trained models with a pre-trained navigation model. We use a visual navigation model (\gcc: ViNG~\cite{shah2020ving}) to create a topological ``mental map'' of the environment using the robot's observations. Given free-form textual instructions, we use a pre-trained large language model (\llm: GPT-3~\cite{brown2020gpt3}) to decode the instructions into a sequence of textual landmarks. We then use a vision-language model (\vlm: CLIP~\cite{radford2021learning}) for \emph{grounding} these textual landmarks in the topological map, by inferring a joint likelihood over the landmarks and nodes. A novel search algorithm is then used to maximize a probabilistic objective, and find a plan for the robot, which is then executed by \gcc.

Our primary contribution is \textbf{L}arge \textbf{M}odel \textbf{Nav}igation, or \sysName, an embodied instruction following system that combines three large independently pre-trained models --- a self-supervised robotic control model that utilizes visual observations and physical actions (\gcc), a vision-language model that grounds images in text but has no context of embodiment (\vlm), and a large language model that can parse and translate text but has no sense of visual grounding or embodiment (\llm) --- to enable long-horizon instruction following in complex, real-world environments. \emph{We present the first instantiation of a robotic system that combines the confluence of pre-trained vision-and-language models with a goal-conditioned controller, to derive actionable plans \emph{without any fine-tuning} in the target environment.} Notably, all three models are trained on large-scale datasets, with self-supervised objectives, and used off-the-shelf with \emph{no fine-tuning} --- no human annotations of the robot navigation data are necessary to train \sysName. We show that \sysName is able to successfully follow natural language instructions in new environments over the course of 100s of meters of complex, suburban navigation, while disambiguating paths with fine-grained commands.

\section{Related Work}
\label{sec:related_work}

Early works in augmenting navigation policies with natural language commands use statistical machine translation~\cite{koehn2009} to discover data-driven patterns to map free-form commands to a formal language defined by a grammar~\cite{wong2006semantic, matuszek2010directions, chen2011interpret, tellex2011understanding, matuszek2013parse}. However, these approaches tend to operate on structured state spaces. Our work is closely inspired by methods that instead reduce this task to a sequence prediction problem~\cite{shimizu2009follow, mei2016listen, anderson2018vision}. Notably, our goal is similar to the task of VLN --- leveraging fine-grained instructions to control a mobile robot solely from visual observations~\cite{anderson2018vision, gu2022vision}.

However, most recent approaches to VLN use a large dataset of simulated trajectories --- over 1M demonstrations --- annotated with fine-grained language labels in indoor~\cite{ku2020rxr, jain2019stay, yan2019xl, anderson2018vision, shridhar2020alfred} and driving scenarios~\cite{chen2019touchdown, hermann2020street, mirowski2019street, vasudevan2021talk, misra2018lani, blukis2018unity}, and rely on sim-to-real transfer for deployment in simple indoor environments~\cite{krantz2020vlnce, anderson2021sim}. However, this necessitates building a photo-realistic simulator resembling the target environment, which can be challenging for unstructured environments, especially for the task of outdoor navigation. Instead, \sysName leverages free-form textual instructions to navigate a robot in complex, outdoor environments \emph{without} access to any simulation or any trajectory-level annotations.

Recent progress in using large-scale models of natural language and images trained on diverse data has enabled applications in a wide variety of textual~\cite{wolf2019huggingface, thoppilan2022lamda, chen2021codex}, visual~\cite{radford2021learning, ramesh2022dalle2, saharia2022imagen, gu2022openvocabulary, chao2021alttext, song2022clip}, and embodied domains~\cite{shridhar2021cliport, jang2021bcz, huang2022language, ahn2022saycan, zeng2022socraticmodels, khandelwal2021clip}.
In the latter category, \citet{shridhar2021cliport}, \citet{khandelwal2021clip} and \citet{jang2021bcz} fine-tune embeddings from pre-trained models on robot data with language labels, \citet{huang2022language} assume that the low-level agent can execute textual instructions (without addressing control), and \citet{ahn2022saycan} assumes that the robot has a set of text-conditioned skills that can follow atomic textual commands. All of these approaches require access to low-level skills that can follow rudimentary textual commands, which in turn requires language annotations for robotic experience and a strong assumption on the robot's capabilities.
In contrast, we combine these pre-trained vision and language models with pre-trained visual policies that do not use any language annotations~\cite{shah2020ving, shah2021rapid} \emph{without} fine-tuning these models in the target environment or for the task of VLN.

Data-driven approaches to vision-based mobile robot navigation often use photorealistic simulators~\cite{savva2019habitat, xiazamirhe2018gibsonenv, savva2017minos, kolve2017thor} or supervised data collection ~\cite{francis2020longrange} to learn goal-reaching policies directly from raw observations. Self-supervised methods for navigation~\cite{Manderson2010SelfSupervisedGP, boris2006improving, gandhi2017fly, kouris2018fly, kahn2018gcg, hirose2019deep, shah2020ving}
instead can use unlabeled datasets of trajectories by automatically generating labels using onboard sensors and hindsight relabeling. Notably, such a policy can be trained on large, diverse datasets and generalize to previously unseen environments~\cite{shah2021rapid, shah2022viking}. Being self-supervised, such policies are adept at navigating to desired goals specified by GPS locations or images, but are unable to parse high-level instructions such as free-form text. \sysName uses self-supervised policies trained in a large number of prior environments, augmented with pre-trained vision and language models for parsing natural language instructions, and deploys them in novel real-world environments \emph{without} any fine-tuning.

\section{Preliminaries}
\label{sec:preliminaries}

\begin{wrapfigure}{R}{0.6\textwidth}
    \centering
    \includegraphics[width=\linewidth]{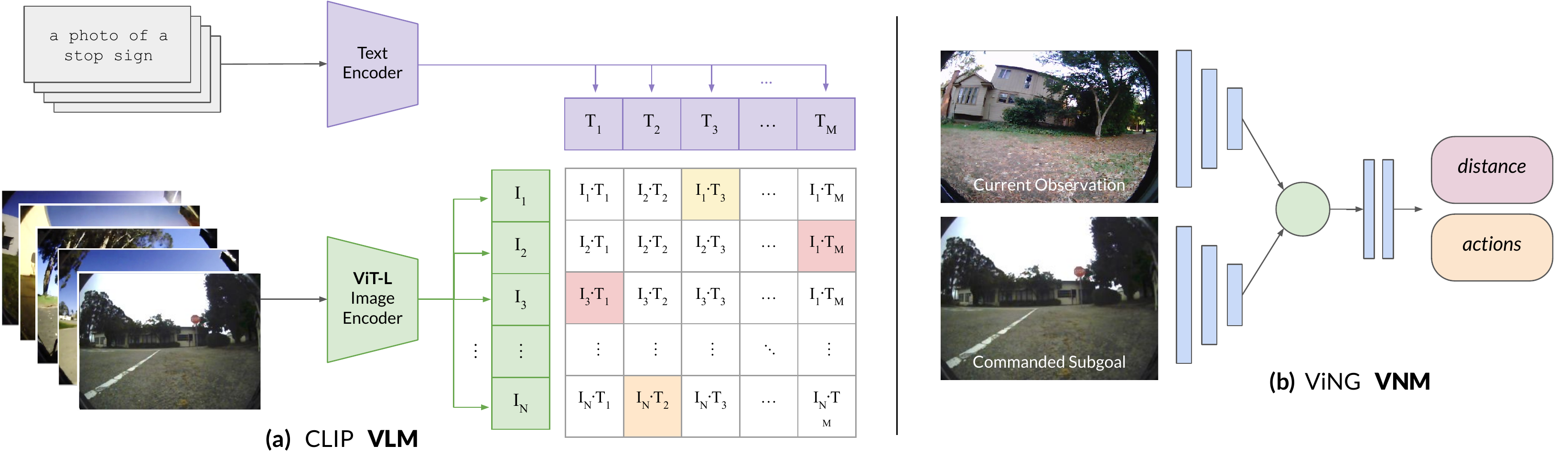}
    \caption{\sysName uses \vlm to infer a joint probability distribution over textual landmarks and image observations. \gcc constitutes an image-conditioned distance function and policy that can control the robot.}
    \label{fig:models}
\end{wrapfigure}

\sysName consists of three large, pre-trained models for processing language, associating images with language, and visual navigation.

\noindent \textbf{Large language models} are generative models based on the Transformer architecture~\cite{vaswani2017attention}, trained on large corpora of internet text. \sysName uses the GPT-3 \llm~\cite{brown2020gpt3}, to parse textual instructions into a sequence of landmarks.

\begin{figure}[t]
    \centering
    \includegraphics[width=0.95\linewidth]{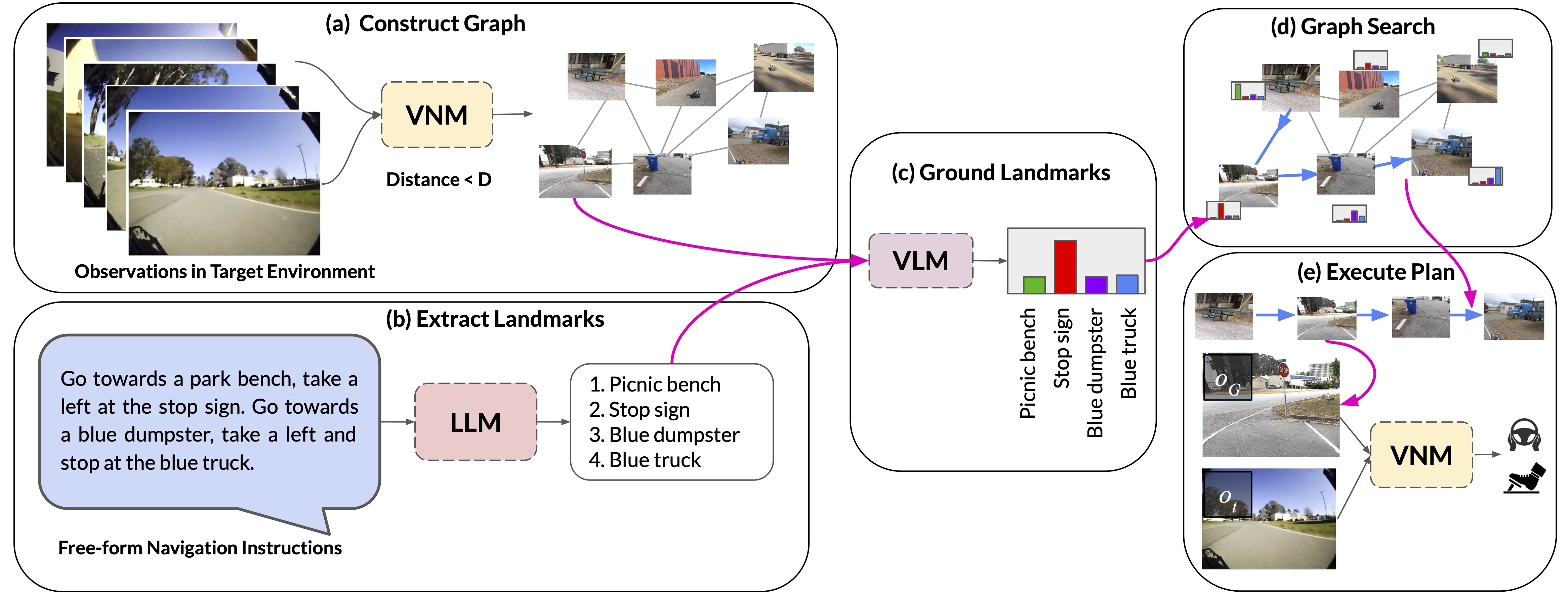}
    \caption{\textbf{System overview:} (a) \gcc uses a goal-conditioned distance function to infer connectivity between the set of raw observations and constructs a topological graph. (b) \llm translates natural language instructions into a sequence of textual landmarks. (c) \vlm infers a joint probability distribution over the landmark descriptions and nodes in the graph, which is used by (d) a graph search algorithm to derive the optimal walk through the graph. (e) The robot drives following the walk in the real world using the \gcc policy.}
    \label{fig:components}
\end{figure}

\noindent \textbf{Vision-and-language models} refer to models that can associate images and text, e.g. image captioning, visual question-answering, etc.~\cite{alayrac2022flamingo, li2019visualbert, chen2020uniter}. We use the CLIP \vlm~\cite{radford2021learning}, a model that jointly encodes images and text into an embedding space that allows it to determine how likely some string is to be associated with a given image. We can jointly encode a set of landmark descriptions $t$ obtained from the \llm and a set of images $i_k$ to obtain their \vlm embeddings $\{T, I_k\}$ (see Fig.~\ref{fig:components}). Computing the cosine similarity between these embeddings, followed by a softmax operation results in probabilities $P(i_k | t)$, corresponding to the likelihood that image $i_k$ corresponds to the string $t$. \sysName uses this probability to align landmark descriptions with images.

\noindent \textbf{Visual navigation models} learn navigation behavior and navigational affordances directly from visual observations~\cite{savinov2018sptm, chaplot2020ans, hirose2019deep, wijmans2020ddppo, shah2020ving}, associating images and actions through time. We use the ViNG \gcc~\cite{shah2020ving}, a goal-conditioned model that predicts temporal distances between pairs of images and the corresponding actions to execute (see Fig.~\ref{fig:components}). This provides an interface between images and embodiment. The \gcc  serves two purposes: (i) given a set of observations in the target environment, the distance predictions from the \gcc can be used to construct a topological graph $\gG(V, E)$ that represents a ``mental map'' of the environment; (ii) given a ``walk'', comprising of a sequence of connected subgoals to a goal node, the \gcc can navigate the robot along this plan. The topological graph $\gG$ is an important abstraction that allows a simple interface for planning over past experience in the environment and has been successfully used in prior work to perform long-horizon navigation~\cite{shah2022viking, meng2020scaling, bruce2018deployable}. To deduce connectivity in $\gG$, we use a combination of learned distance estimates, temporal proximity (during data collection), and spatial proximity (using GPS measurements). For every connected pair of vertices $\{v_i, v_j\}$, we assign this distance estimate to the corresponding edge weight $D(v_i, v_j)$. For more details on the construction of this graph, see Appendix~\ref{app:graph}.

\section{\sysName: Instruction Following with Pre-Trained Models}
\label{sec:method}

\sysName combines the components discussed earlier to follow textual instructions in the real world. The \llm parses free-form instructions into a list of landmarks $\bar{\ell}$ (Sec.~\ref{sec:text2nav}), the \vlm associates these landmarks with nodes in the graph by estimating the probability that each node $\bar{v}$ corresponds to each $\bar{\ell}$, $P (\bar{v} | \bar{\ell})$ (Sec.~\ref{sec:landmark2loc}), and the \gcc is then used to infer how effectively the robot can navigate between each pair of nodes in the graph, which we convert into a probability $P(\overline{v_i, v_j})$ derived from the estimated temporal distances.
To find the optimal ``walk'' on the graph that both (i) adheres to the provided instructions and (ii) minimizes traversal cost, we derive a probabilistic objective (Sec.~\ref{sec:problem}) and show how it can be optimized using a graph search algorithm (Sec.~\ref{sec:graph_search}). This optimal walk is then executed in the real world by using the actions produced by the \gcc model.

\subsection{Problem Formulation}
\label{sec:problem}

Given a sequence of landmark descriptions $\bar{\ell} = \ell_1, \ell_2, ..., \ell_n$ extracted by the \llm from the language command, our method needs to determine a sequence of waypoints $\bar{v} = v_1, v_2, ..., v_k$ to command to the robot (typically, $k \geq n$, since each landmark needs to be visited, but the traversal might also require traversing other waypoints in-between the landmarks). We can formulate this as a probabilistic inference problem. A key element in this formulation is access to a distribution $p(v_i | \ell_j)$ for each graph vertex $v_i$ and landmark description $\ell_j$. Recall that the graph vertices correspond to images observed by the robot, and thus, $p(v_i | \ell_i)$ represents a distribution over images given a language description. This can be obtained directly from the \vlm. Intuitively, the full likelihood that we need to optimize to determine the robot's plan will now depend on two terms: likelihoods of the form $p(v_{t_i} | \ell_i)$ that describe how likely $v_{t_i}$ is to correspond to $\ell_i$ for an assignment $t_1, t_2, \ldots, t_n$, and traversability likelihoods $p(\overline{v_i, v_{i+1}})$ that describe how likely is the robot to be able to reach $v_{i+1}$ from $v_i$, which can be determined via the \gcc. 

While we can use a variety of traversability likelihood functions, a simple choice is to use a discounted Markovian model, where the discount $\gamma$ models the probability of \emph{exiting} at each time step, leading to a termination probability of $1-\gamma$ at each step, and a probability of reaching $v_{i+1}$ given by $\gamma^{D(v_i, v_{i+1})}$, where $D(v_i, v_{i+1})$ is the estimated number of time steps the robot needs to travel from $v_i$ to $v_{i+1}$, which is predicted by the \gcc. While other traversability likelihoods could also be used, this choice is a convenient consequence of goal-conditioned reinforcement learning formulations~\citep{kaelbling1993learning, hartikainen2020}, and thus, the log-likelihood corresponds to $D(v_i, v_{i+1})$. We can use these likelihoods to derive the probability that a given sequence $\bar{v}$ can be traversed successfully, which we denote with the auxiliary Bernoulli random variable $c_{\bar{v}}$ (i.e., $c_{\bar{v}} = 1$ implies that $\bar{v}$ was traversed successfully):
\begin{equation}
    P(c_{\bar{v}} = 1 | \bar{v}) = \prod_{1 \le i < T} P(\overline{v_i, v_{i+1}}) = \prod_{1 \le i < T} {\gamma}^{D(v_i, v_{i+1})},
    \label{eq:def:p_t}
\end{equation}
The full likelihood used for planning is then given by:%
\begin{equation}
    P(\text{success} | \bar{v}, \bar{\ell}) \propto P(c_{\bar{v}} = 1 | \bar{v}) P(\bar{v} | \bar{\ell})\; 
    = \prod_{1 \le j < k} {\gamma}^{D(v_j, v_{j+1})} \max_{1 \le t_1 \le \ldots \le t_n \le k} \prod_{1 \le i \le n} P(v_{t_i} | \ell_i).
    \label{eq:def:p_objective}
\end{equation}

\subsection{Parsing Free-Form Textual Instructions}
\label{sec:text2nav}

The user specifies the route they want the robot to take  using natural language, while the objective above is defined in terms of a sequence of desired landmarks. To extract this sequence from the user's natural language instruction we employ a standard large language model, which in our prototype is GPT-3 \cite{brown2020gpt3}. We used a prompt with 3 examples of correct landmarks' extractions, followed up by the description to be translated by the \llm. Such an approach worked for the instructions that we tested it on. Examples of instructions together with landmarks extracted by the model can be found in Fig.~\ref{fig:qualitative}. The appropriate selection of the prompt, including those 3 examples, was required for more nuanced cases. For details of the \emph{``prompt engineering''} please see Appendix~\ref{app:prompt_engineering}.

\subsection{Visually Grounding Landmark Descriptions}
\label{sec:landmark2loc}

\begin{wrapfigure}[15]{R}{0.55\textwidth}
\vspace{-1.5em}
\begin{minipage}{0.54\textwidth}
\begin{algorithm}[H]
\caption{Graph Search}
\begin{algorithmic}[1]
\STATE \textbf{Input}: Landmarks $(\ell_1, \ell_2, \ldots, \ell_n)$.
\STATE \textbf{Input}: Graph $\gG(V, E)$.
\STATE \textbf{Input}: Starting node $S$.
\STATE $\mathlarger{\mathlarger{\forall_{\substack{i = 0, \ldots, n \\ v \in V}}}} ~~ Q[i, v] = -\infty$
\STATE $Q[0, S] = 0$
\STATE $\text{Dijkstra\_algorithm}(\gG, Q[0, *])$
\FOR{$i$ in $1, 2, \ldots, n$}
\STATE $\mathlarger{\mathlarger{\forall{v \in V}}} Q[i, v] = Q[i-1, v] + \textrm{CLIP}(v, \ell_i)$
\STATE Dijkstra\_algorithm($\gG, Q[i, *]$)
\ENDFOR
\STATE $\text{destination} = \mathrm{arg}\max(Q[n, *])$
\STATE return $\mathrm{backtrack}(\text{destination}, Q[n, *])$
\end{algorithmic}
\label{alg:dynamic_programming}
\end{algorithm}
\end{minipage}
\end{wrapfigure}

As discussed in Sec.~\ref{sec:problem}, a crucial element of selecting the walk through the graph is computing $P(v_i | \ell_j)$, the probability that landmark description $v_i$ refers to node $\ell_j$ (see Eqn.~\ref{eq:def:p_objective}). With each node containing an image taken during initial data collection, the probability can be computed using CLIP \cite{radford2021learning} in the way described in Sec.~\ref{sec:preliminaries} as the retrieval task. As presented in Fig.~\ref{fig:models}, we apply CLIP to the image at node $v_i$ and caption prompt in the form of \textit{``This is a photo of a [$\ell_j$]''}.
To go from CLIP model outputs, which are logits, to probabilities we use $P(v_i | \ell_j) = \frac{\exp \text{CLIP}(v_i, \ell_j)}{\sum_{v \in V} \exp \text{CLIP}(v, \ell_j)}$.
The resulting probability $P(v_i | \ell_j)$, together with the inferred edges' distances will be used to select the optimal walk.

\subsection{Graph Search for the Optimal Walk}
\label{sec:graph_search}

As described in Sec.~\ref{sec:problem}, \sysName aims at finding a walk $\bar{v} = (v_1, v_2, \ldots, v_k)$ that maximizes the probability of successful execution of $\bar{v}$ that adheres to the given list of landmarks $\bar{\ell}$. We can define a function $R(\bar{v}, \bar{t})$ for a monotonically increasing sequence of indices $\bar{t} = (t_1, t_2, \ldots, t_n)$:
\begin{equation}
  R(\bar{v}, \bar{t}) := \sum_{i=1}^{n} \text{CLIP}(v_{t_i}, \ell_i) - \alpha \sum_{j=1}^{T-1} D(v_j, v_{j+1}), \text{where}\, \alpha = - \log \gamma.
  \label{eq:score_definition}
\end{equation}
$R$ has the property that $(\bar{v})$ maximizes $P(\text{success} | \bar{v}, \bar{\ell})$ defined in Eqn.~\ref{eq:def:p_objective}, if and only if there exists $\bar{t}$ such that $(\bar{v}, \bar{t})$ maximizes $R$.
In order to find such $(\bar{v}$, $\bar{t})$, we employ dynamic programming. In particular we define a helper function $Q(i, v)$ for $i \in \{ 0, 1, \ldots, n\}$, $v \in V$:
\begin{equation}
    Q(i, v) = \max_{\substack{\bar{v} = (v_1, v_2, \ldots, v_j), v_j = v \\ \bar{t} = (t_1, t_2, \ldots, t_i)}} R(\bar{v}, \bar{t}).
    \label{eq:q_definition}
\end{equation}
$Q(i, v)$ represents the maximal value of $R$ for a walk ending in $v$ that visited the landmarks up to index $i$. The base case $Q(0,v)$ visits none of the landmarks, and its value of $R$ is simply equal to minus the length of shortest path from the starting node $S$. For $i > 0$ we have:
\begin{equation}
    Q(i, v) = \max \bigg (
        Q(i-1, v) + \text{CLIP} (v, \ell_i),
        \max_{w \in \text{neighbors}(v)} Q(i, w) - \alpha \cdot D(v, w)
    \bigg ).
    \label{eq:q_update_rule}
\end{equation}
The base case for DP is to compute $Q(0, V)$. Then, in each step of DP $i = 1, 2, \ldots, n$ we compute $Q(i, v)$. This computation resembles the Dijkstra algorithm (\cite{dijkstra1959note}). In each iteration, we pick the node $v$ with the largest value of $Q(i, v)$ and update its neighbors based on the Eqn.~\ref{eq:q_update_rule}. Algorithm~\ref{alg:dynamic_programming} summarizes this search process. The result of this algorithm is a walk $\bar{v} = (v_1, v_2, \ldots, v_k)$ that maximizes the probability of successfully carrying out the instruction. Such a walk can be executed by \gcc, using its action estimates to sequentially navigate to these nodes.

\section{System Evaluation}
\label{sec:results}

We now describe our experiments deploying \sysName in a variety of outdoor settings to follow high-level natural language instructions with a small ground robot. For all experiments, the weights of \llm, \vlm, and \gcp are frozen --- there is \emph{no fine-tuning or annotation} in the target environment. We evaluate the complete system, as well as the individual components of \sysName, to understand its strengths and limitations. Our experiments demonstrate the ability of \sysName to follow high-level instructions, disambiguate paths, and reach goals that are up to 800m away.

\subsection{Mobile Robot Platform}
We implement \sysName on a Clearpath Jackal UGV platform (see Fig.~\ref{fig:overview}(right)).
The sensor suite consists of a 6-DoF IMU, a GPS unit for approximate localization, wheel encoders for local odometry, and front- and rear-facing RGB cameras with a $170^\circ$ field-of-view for capturing visual observations and localization in the topological graph. The \llm and \vlm queries are pre-computed on a remote workstation and the computed path is commanded to the robot wirelessly. The \gcc runs on-board and only uses forward RGB images and unfiltered GPS measurements.

\subsection{Following Instructions with \sysName}
\label{sec:experiments}
\begin{figure}[t]
    \centering
    \includegraphics[width=\textwidth]{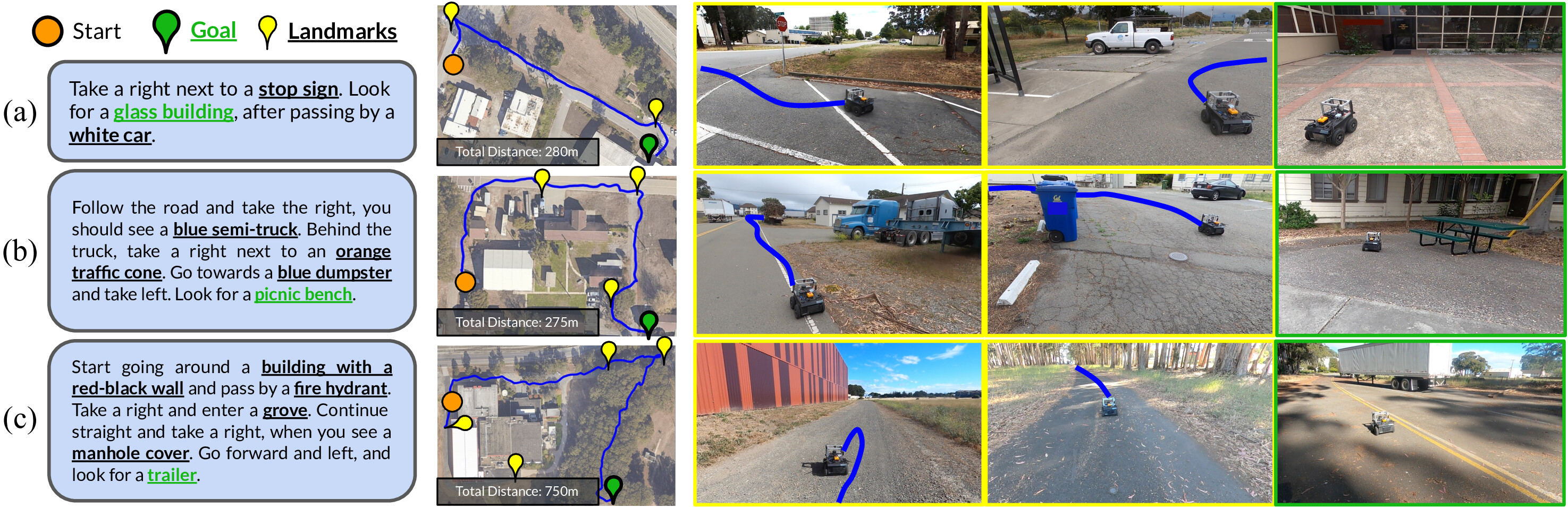}
    \caption{\textbf{Qualitative examples} of \sysName in real-world environments executing textual instructions (left). The landmarks extracted by \llm (highlighted in text) are grounded into visual observations by \vlm (center; overhead image not available to the robot). The resulting \emph{walk} of the graph is executed by \gcp (right).}
    \label{fig:qualitative}
    \vspace*{-1em}
\end{figure}

In each evaluation environment, we first construct the graph by manually driving the robot and collecting image and GPS observations. The graph is constructed automatically using the \gcc from this data, and in principle such data could also be obtained from past traversals, or even with autonomous exploration methods~\citep{shah2021rapid}. Once the graph is constructed, the robot can carry out instructions in that environment.
We tested our system on 20 queries, in environments of varying difficulty, corresponding to a total combined length of over 6 km. Instructions include a set of prominent landmarks in the environment that can be identified from the robot’s observations, e.g. traffic cones, buildings, stop signs, etc.

Fig.~\ref{fig:qualitative} shows qualitative examples of the path taken by the robot. Note that the overhead image and spatial localization of the landmarks is \emph{not} available to the robot and is shown for visualization only.
In Fig.~\ref{fig:qualitative}(a), \sysName is able to successfully localize the simple landmarks from its prior traversal and find a short path to the goal. While there are multiple stop signs in the environment, the objective in Eqn.~\ref{eq:def:p_objective} causes the robot to pick the correct stop sign in context, so as to minimize overall travel distance. Fig.~\ref{fig:qualitative}(b) highlights \sysName's ability to parse complex instructions with multiple landmarks specifying the route --- despite the possibility of a shorter route directly to the final landmark that ignores instructions, the robot finds a path that visits all of the landmarks in the correct order.

\begin{wrapfigure}{R}{0.48\columnwidth}
    \centering
    \vspace*{-0.1em}
    \includegraphics[width=0.48\columnwidth]{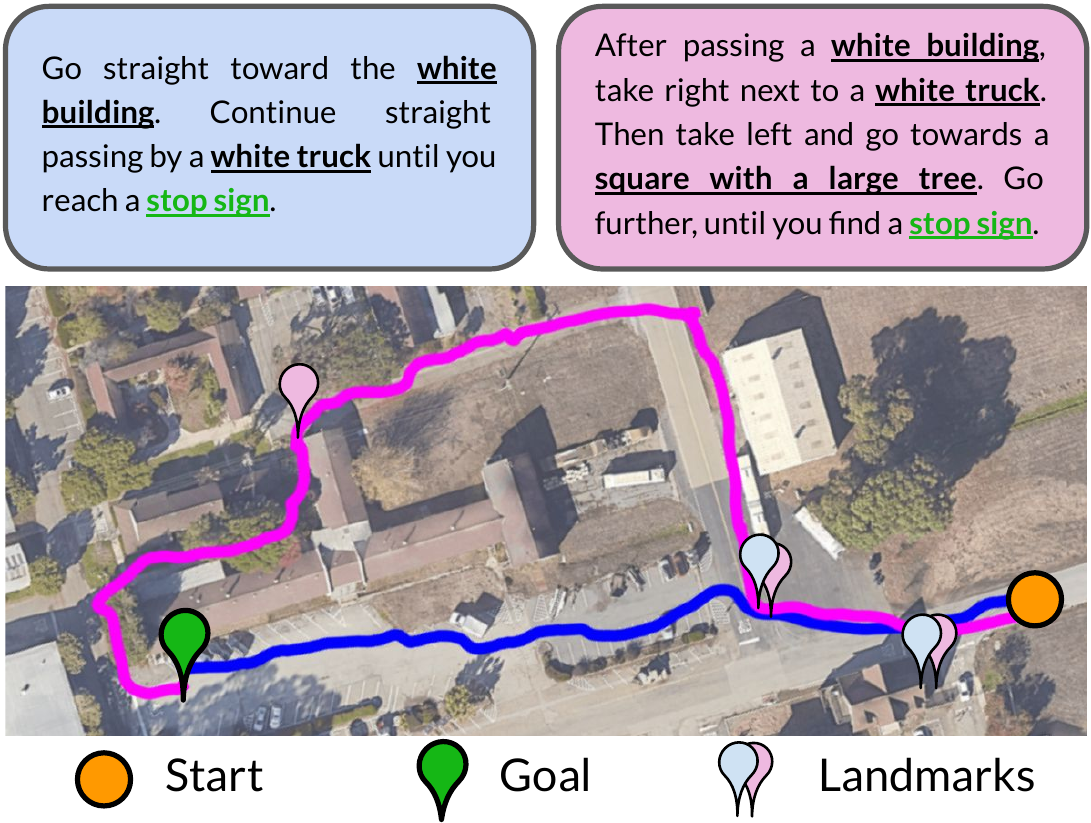}
    \caption{\sysName can successfully disambiguate instructions with same {\color{orange}start}-{\color{green}goal} locations that differ slightly, and execute them. Extracted landmarks and their corresponding locations are highlighted and marked with a pin, respectively.}
    \label{fig:disambiguation}
    \vspace*{-1.1em}
\end{wrapfigure}

\noindent\textbf{Disambiguation with instructions.} Since the objective of \sysName is to follow instructions, and not merely to reach the final goal, different instructions may lead to different traversals. Fig.~\ref{fig:disambiguation} shows an example where modifying the instruction can disambiguate multiple paths to the goal. Given the shorter prompt (blue), \sysName prefers the more direct path. On specifying a more fine-grained route (magenta), \sysName takes an alternate path that passes a different set of landmarks.

\noindent\textbf{Missing landmarks.} While \sysName is effective at parsing landmarks from instructions, localizing them on the graph, and finding a path to the goal, it relies on the assumption that the landmarks (i) exist in the environment, and (ii) can be identified by the \vlm. Fig.~\ref{fig:qualitative}(c) illustrates a case where the executed path fails to visit one of the landmarks --- a fire hydrant --- and takes a path that goes around the top of the building rather than the bottom. This failure mode is attributed to the the inability of the \vlm to detect a fire hydrant from the robot's observations. On independently evaluating the efficacy of our the \vlm at retrieving landmarks (see Sec.~\ref{sec:ablations}), we find that despite being the best off-the-shelf model for our task, CLIP is unable to retrieve a small number of ``hard'' landmarks, including fire hydrants and cement mixers. In many practical cases, the robot is still successful in finding a path that visits the remaining landmarks.

\subsection{Quantitative Analysis}
\label{sec:quantitative_analysis}

To quantify the performance of \sysName, we introduce some performance metrics. A walk produced by the graph search is considered \emph{successful}, if (1) it matches the path intended by the user or (2) if the landmark images extracted by the search algorithm indeed contain said landmarks (i.e. if the produced path is \emph{valid}, if not identical). The fraction of successful walks produced by the search algorithm is defined as \emph{planning success}. For a successfully executed plan in the real world, we define \emph{efficiency} as: \[\min(1, \frac{\text{length of described route}}{ \text{length of executed route}})\] The second term --- corresponding to the optimality of the executed route --- is clipped at a maximum of 1 to account for occasional cases when the \gcc executes a shorter, more direct path than the user intended. For a set of queries, we report the average efficiency over successful experiments. The \emph{planning efficiency} is analogously defined as: \[\min(1, \frac{\text{length of described route}}{ \text{length of planned walk}})\] Yet another metric --- \emph{number of disengagements} --- counts the average number of human interventions required per experiment, due to unsafe maneuvers like collisions or falling off a curb, etc.

Table~\ref{tab:master_table} summarizes the quantitative performance of the system over 20 instructions. \sysName can consistently follow the instructions in 85\% of the experiments, without collisions or disengagements (an average of 1 intervention per 6.4km of traversals). Comparing to baselines where the navigation model has been ablated (described in Sec.~\ref{sec:ablations}), \sysName performs consistently better in executing efficient, collision-free paths to the goal. In all the unsuccessful experiments, the failure can be attributed to the inability of the planning stage --- the search algorithm is unable to visually localize certain ``hard'' landmarks in the graph --- leading to incomplete execution of the instructions. Investigating these failure modes suggests that the most critical component of our system is the ability of \vlm to detect unfamiliar landmarks, e.g. a fire hydrant, and in challenging lighting conditions, e.g. underexposed images.

\begin{table}
\centering
{\footnotesize
\begin{tabular}{ll|c|cccc}
\toprule
System & Environment & Net Success $\uparrow$ & Efficiency $\uparrow$ & \# Diseng. $\downarrow$ & Planning $\uparrow$ \\ \midrule
GPS-Nav (No \gcc) & \texttt{EnvSmall-10} & 0.23 & 0.93 & 0.75 & \textbf{0.9} \\ \midrule
\rowcolor{Cerulean!20}  & \texttt{EnvSmall-10} & \textbf{0.8} & \textbf{0.96} & \textbf{0.1} &\textbf{ 0.9}\\
\rowcolor{Cerulean!20} \multirow{-2}{*}{{\sysName (Ours)}} & \texttt{EnvLarge-10} & \textbf{0.8} & \textbf{0.89} & \textbf{0} & \textbf{0.8}\\
\bottomrule \\
\end{tabular}}
\caption{Quantifying navigational instruction following with \sysName over 20 experiments. \sysName can successfully plan a path to the goal, and follow it efficiently, over 100s of meters. Ablating the \gcc (GPS-Nav) severely hurts performance due to frequent disengagements inability to reason about collisions with obstacles.}
\label{tab:master_table}
\vspace*{-1em}
\end{table}

\subsection{Dissecting \sysName}
\label{sec:ablations}

To understand the influence of each of the components of \sysName, we conduct experiments to evaluate these components in isolation. For more details about these experiments, see Appendix~\ref{app:ablations}.

\begin{figure}[t!]
    \centering
    \begin{minipage}{0.50\textwidth}
        \centering
        {\footnotesize
        \begin{tabular}{l c}
            \toprule
           \llm Candidate & Avg. Extraction Success\\ 
            \midrule
        Noun Chunks   & 0.88 \\
        fairseq-1.3B~\cite{artetxe2021efficient}       & 0.52    \\
        fairseq-13B~\cite{artetxe2021efficient}       & 0.76    \\
        GPT-J-6B~\cite{gpt-j}       & 0.80    \\
        GPT-NeoX-20B~\cite{black2022gpt_neox_20b}       & 0.72    \\
        \rowcolor{Cerulean!20}{GPT-3~\cite{brown2020gpt3}}             & \textbf{1.0}   \\
        \bottomrule \\
        \end{tabular}}
        \captionof{table}{GPT-3 consistently outperforms alternatives in parsing free-form instructions into landmarks.}
        \label{tab:llm_ablate}
    \end{minipage}
    ~~
    \begin{minipage}{0.45\textwidth}
        \centering
        {\footnotesize
        \begin{tabular}{l c}
            \toprule
           \vlm Candidate & Detection Rate\\ 
            \midrule
        Faster-RCNN~\cite{ren2015frcnn}                   & 0.07 \\ %
        ViLD~\cite{gu2022openvocabulary}         & 0.38    \\ %
        \rowcolor{Cerulean!20}{CLIP-ViT~\cite{radford2021learning}}      & \textbf{0.87}    \\ %
        \bottomrule \\
        \end{tabular}}
        \captionof{table}{CLIP-ViT produces the most reliable landmark detections from visual observations.} \label{tab:vlm_ablate}
    \end{minipage}
    \vspace*{-1.4em}
\end{figure}

\begin{table}
\centering
{\footnotesize
\begin{tabular}{lccccc}
\toprule
& \multicolumn{2}{c}{\texttt{EnvSmall-10}} & \multicolumn{2}{c}{\texttt{EnvLarge-10}} \\
Planner & Pl. Success $\uparrow$ & Pl. Efficiency $\uparrow$ & Pl. Success $\uparrow$ & Pl. Efficiency $\uparrow$ \\ \midrule
Max Likelihood & 0.6 & 0.69 & 0.2 & 0.17 \\
\rowcolor{Cerulean!20} \sysName (Ours) & \textbf{0.9} & \textbf{0.80} & \textbf{0.8} & \textbf{0.99} \\
\bottomrule \\
\end{tabular}}
\caption{Ablating the search algorithm (Sec.~\ref{sec:graph_search}) gives a max likelihood planner that ignores reachability information, resulting in inefficient plans that are upto $6\times$ longer than \sysName for the same instruction.}
\label{tab:max_likelihood}
\end{table}

To evaluate the performance of \llm candidates in parsing instructions into an \emph{ordered} list of landmarks, we compare GPT-3 (used by \sysName) to other state-of-the-art pre-trained language models --- fairseq~\cite{artetxe2021efficient}, GPT-J-6B~\cite{gpt-j}, and GPT-NeoX-20B~\cite{black2022gpt_neox_20b} --- as well as a simple baseline using spaCy NLP library~\cite{honnibal2020spacy} that extracts base noun phrases, followed by filtering. In Table~\ref{tab:llm_ablate} we report the average extraction success for all the methods on the $20$ prompts used in Section~\ref{sec:quantitative_analysis}. GPT-3 significantly outperforms other models, owing to its superior representation capabilities and in-context learning~\cite{rong2021context}. The noun chunking performs surprisingly reliably, correctly solving many simple prompts. For further details on these experiments, see Appendix~\ref{app:sec:llm_ablation}.

To evaluate the \vlm's ability to ground these textual landmarks in visual observations, we set up an object detection experiment. Given an unlabeled image from the robot's on-board camera and a set of textual landmarks, the task is to \emph{retrieve} the corresponding label. We run this experiment on a set of 100 images from the environments discussed earlier, and a set of 30 commonly-occurring landmarks. These landmarks are a combination of the landmarks retrieved by the \llm in our experiments from Sec.~\ref{sec:experiments} and manually curated ones. We report the detection successful if any of the top 3 predictions adhere to the contents of the image. We compare the retrieval success of our \vlm (CLIP) with some credible object detection alternatives --- Faster-RCNN-FPN~\cite{ren2015frcnn, lin2017fpn}, a state-of-the-art object detection model pre-trained on MS-COCO~\cite{Lin2014MicrosoftCC, wu2019detectron2}, and ViLD~\cite{gu2022openvocabulary}, an open-vocabulary object detector based on CLIP and Mask-RCNN~\cite{he2017maskrcnn}. To evaluate against the closed-vocabulary baseline, we modify the setup by projecting the landmarks onto the set of MS-COCO class labels. We find that CLIP outperforms baselines by a wide margin, suggesting that its visual model transfers very well to robot observations (see Table~\ref{tab:vlm_ablate}). Despite deriving from CLIP, ViLD struggles with detecting complex landmarks like ``manhole cover'' and ``glass building''. Faster-RCNN is unable to detect common MS-COCO objects like ``traffic light'', ``person'' and ''stop sign'', likely due to the on-board images being out-of-distribution for the model.

\begin{wrapfigure}{R}{0.41\columnwidth}
    \centering
    \vspace*{-1.3em}
    \includegraphics[width=0.41\columnwidth]{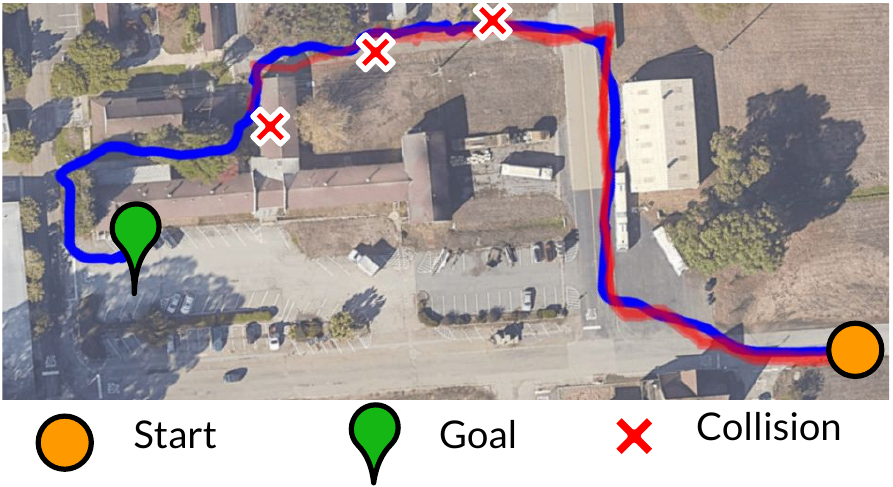}
    \caption{GPS-Nav (red) fails to execute a plan due to its inability to reason about traversability through obstacles, while \sysName (blue) succeeds.}
    \vspace*{-1.3em}
    \label{fig:gps-ablation}
\end{wrapfigure}

To understand the importance of the \gcc, we run an ablation experiment of \sysName without the navigation model. Using GPS-based distance estimates and a na\"{i}ve straight line controller between nodes of the topological graph. Table~\ref{tab:master_table} summarizes these results --- without \gcc's ability to reason about obstacles and traversability, the system frequently runs into small obstacles such as trees and curbs, resulting in failure. Fig.~\ref{fig:gps-ablation} illustrates such a case --- while such a controller works well on open roads, it cannot reason about connectivity around buildings or obstacles, and results in collisions with a curb, a tree, and a wall in 3 individual attempts. This illustrates that using a learned policy and distance function from the \gcc is critical for enabling \sysName to navigate in complex environments without collisions.

Lastly, to understand the importance of the two components of the graph search objective (Eqn.~\ref{eq:score_definition}), we ran a set of ablations where the graph search only depends on $P_l$, i.e. \emph{Max Likelihood Planning}, which only picks the most likely landmark without reasoning about topological connectivity or traversability. Table~\ref{tab:max_likelihood} shows that such a planner suffers greatly in the form of efficiency, because it does not utilize the spatial organization of nodes and their connectivity. For more details on these experiments, and qualitative examples, see Appendix~\ref{app:ablations}.

\vspace*{-0.5em}
\section{Discussion}
\label{sec:discussion}

\vspace*{-0.5em}
We presented \textbf{L}arge \textbf{M}odel \textbf{Nav}igation, or \sysName, a system for robotic navigation from textual instructions that can control a mobile robot without requiring any user annotations for navigational data. \sysName combines three pre-trained models: the \llm, which parses user instructions into a list of landmarks, the \vlm, which estimates the probability that each observation in a ``mental map'' constructed from prior exploration of the environment matches these landmarks, and the \gcc, which estimates navigational affordances (distances between landmarks) and robot actions. Each model is pre-trained on its own dataset, and we show that the complete system can execute a variety of user-specified instructions in real-world outdoor environments --- choosing the correct sequence of landmarks through a combination of language and spatial context --- and handle mistakes (such as missing landmarks). We also analyze the impact of each pre-trained model on the full system.

\noindent \textbf{Limitations and future work.} The most prominent limitation of \sysName is its reliance on landmarks: while the user can specify any instruction they want, \sysName only focuses on the landmarks and disregards any verbs or other commands (e.g., ``go straight for three blocks'' or ``drive past the dog very slowly''). Grounding verbs and other nuanced commands is an important direction for future work. Additionally, \sysName uses a \gcc that is specific to outdoor navigation with the Clearpath Jackal robot. An exciting direction for future work would be to design a more general ``large navigation model'' that can be utilized broadly on any robot, analogous to how the \llm and \vlm handle any text or image. However, we believe that in its current form, \sysName provides a simple and attractive prototype for how pre-trained models can be combined to solve complex robotic tasks, and illustrates that these models can serve as an ``interface'' to robotic controllers that are trained without any language annotations. One of the implications of this result is that further progress on self-supervised robotic policies (e.g., goal-conditioned policies) can directly benefit instruction following systems. More broadly, understanding how modern pre-trained models enable effective decomposition of robotic control may enable broadly generalizable systems in the future, and we hope that our work will serve as a step in this direction.

\vspace*{-0.5em}
\acknowledgments{This research was supported by DARPA Assured Autonomy, DARPA RACER, Toyota Research Institute, ARL DCIST CRA W911NF-17-2-0181, and AFOSR. BO was supported by the Fulbright Junior Research Award granted by the Polish-U.S. Fulbright Commission. We would like to thank Alexander Toshev for pivotal discussions in early stages of the project. We would also like to thank Kuan Fang, Siddharth Karamcheti, and Albertyna Osińska for useful discussions and feedback.}

\bibliography{references}  %

\newpage
\appendix

\part*{Appendix}

\section{Prompt Engineering}
\label{app:prompt_engineering}

To use large language models for a particular task, as opposed to a general text completion, one needs to encode the task as a part of the text input to the model. There exist many ways to create such encoding and the process of the representation optimization is sometimes referred to as \emph{prompt engineering} \cite{radford2021learning}. In this section, we discuss the prompts we used for \llm and \vlm.

\subsection{\llm Prompt Engineering}

\sethlcolor{green}

All our experiments use GPT-3 \cite{brown2020gpt3} as the \llm, accessible via OpenAI's API: \texttt{https://openai.com/api/}. We used this model to extract a list of landmarks from free-form instructions. The model outputs were very reliable and robust to small changes in the input prompts. For parsing simple queries, GPT-3 was surprisingly effective with a single, zero-shot prompt. See the example below, where the model output is highlighted:

\begin{quote}
\texttt{First, you need to find a stop sign. Then take left and right and continue until you reach a square with a tree. Continue first straight, then right, until you find a white truck. The final destination is a white building. \\
Landmarks: \\
1.\hl{ Stop sign \\
2. Square with a tree \\
3. White truck \\
4. White building \\}
}
\end{quote}

While this prompt is sufficient for simple instructions, more complex instructions require the model to reason about occurrences such as re-orderings, e.g. \emph{Look for a glass building after after you pass by a white car}. We leverage GPT-3 ability to perform \emph{in-context learning}~\cite{rong2021context} by adding three examples in the prompt:

\begin{quote}
\texttt{Look for a library, after taking a right turn next to a statue.\\
Landmarks:\\
1. a statue\\
2. a library\\}
\\
\texttt{Look for a statue. Then look for a library. Then go towards a pink house.\\
Landmarks:\\
1. a statue\\
2. a library\\
3. a pink house\\}
\\
\texttt{
[Instructions] \\
Landmarks:\\
1.\hl{~~~}}
\end{quote}

We use the above prompt in all our experiments (Section~\ref{sec:experiments},  \ref{sec:quantitative_analysis}), and GPT-3 was successfully able to extract all landmarks. The comparison to other extraction methods is described in Section~\ref{sec:ablations} and Appendix~\ref{app:sec:llm_ablation}.

\subsection{\vlm Prompt Engineering}
In the case of our \vlm --- CLIP \cite{radford2021learning} --- we use a simple family of prompts: \emph{This is a photo of \_\_\_}, appended with the landmark description. This simple prompt was sufficient to detect over $95\%$ of the landmarks encountered in our experiments. While our experiments did not require more careful prompt engineering, \citet{radford2021learning} and \citet{zeng2022socraticmodels} report improved robustness by using an ensemble of slightly varying prompts.

\section{Building the Topological Graph with \gcc}
\label{app:graph}
This section outlines finer details regarding how the topological graph is constructed using \gcc. We use a combination of learned distance estimates (from \gcc), spatial proximity (from GPS), and temporal proximity (during data collection), to deduce edge connectivity. If the corresponding timestamps of two nodes are close ($<2s$), suggesting that they were captured in quick succession, then the corresponding nodes are connected --- adding edges that were physically traversed. If the \gcc estimates of the images at two nodes are close, suggesting that they are \emph{reachable}, then the corresponding nodes are also connected --- adding edges between distant nodes along the same route and giving us a mechanism to connect nodes that were collected in different trajectories or at different times of day but correspond to the nearby locations. To avoid cases of underestimated distances by the model due to aliased observations, e.g. green open fields or a white wall, we filter out prospective edges that are significantly further away as per their GPS estimates --- thus, if two nodes are nearby as per their GPS, e.g. nodes on different sides of a wall, they may not be disconnected if the \gcc does not estimate a small distance; but two similar-looking nodes 100s of meters away, that may be facing a white wall, may have a small \gcc estimate but are not added to the graph to avoid \emph{wormholes}. Algorithm~\ref{alg:graph_building} summarizes this process --- the timestamp threshold $\epsilon$ is 1 second, the learned distance threshold $\tau$ is 80 time steps (corresponding to $\sim 20$ meters), and the spatial threshold $\eta$ is 100 meters.

\begin{algorithm}
\caption{Graph Building}
\begin{algorithmic}[1]
\label{alg:graph_building}
\STATE \textbf{Input}: Nodes $n_i, n_j \in \gG$ containing robot observations; \gcc distance function $f_d$; hyperparameters $\{\tau, \epsilon, \eta\}$
\STATE \textbf{Output}: Boolean $e_{ij}$ corresponding to the existence of edge in $\gG$, and its weight
\STATE $\text{learned distance}\; D_{ij} = f_d(n_i[\text{`image'}], n_j[`\text{image}'])$
\STATE $\text{timestamp distance}\; T_{ij} = \lvert n_i[\text{`timestamp'}] - n_j[\text{`timestamp'}] \rvert$
\STATE $\text{spatial distance}\; X_{ij} = \| n_i[\text{`GPS'}] - n_j[\text{`GPS'}]) \|$
\STATE \textbf{if} ( $ T_{ij} < \epsilon$) \textbf{then} return \textit{\{True, $D_{ij}$\}}
\STATE \textbf{else if} ($D_{ij} < \tau$)  AND  ($ X_{ij}<\eta$) \textbf{then} return \textit{\{True, $D_{ij}$\}}
\STATE \textbf{else} return \textit{False}
\end{algorithmic}
\end{algorithm}
Since a graph obtained by such an analysis may be quite dense, we perform a \emph{transitive reduction} operation on the graph to remove redundant edges.

\section{Miscellaneous Ablation Experiments}
\label{app:ablations}

\subsection{Ablating the Search Objective}

The graph search objective described in Section~\ref{sec:graph_search} can be factored into two components: visiting the required landmarks (denoted by $P_{l} (\bar{v} | \bar{l})$) and minimizing distance traveled (denoted by $P_{t} (\bar{v})$). To analyze the importance of these two components, we ran a set of experiments where the nodes to be visited are selected based only on $P_{l}$. This corresponds to a \textit{Max Likelihood} planner, which only picks the most likely node for each landmark, without reasoning about their relative topological positions and traversability. This approach leads to a simpler algorithm: for each of the landmark descriptions, the algorithm selects the node with the highest CLIP score and connects it via the shortest path to the current node. The shortest path between each pair of nodes is computed using the Floyd–Warshall algorithm.

\begin{figure}[t]
    \begin{subfigure}{.5\textwidth}
        \centering
        \includegraphics[width=0.9\textwidth]{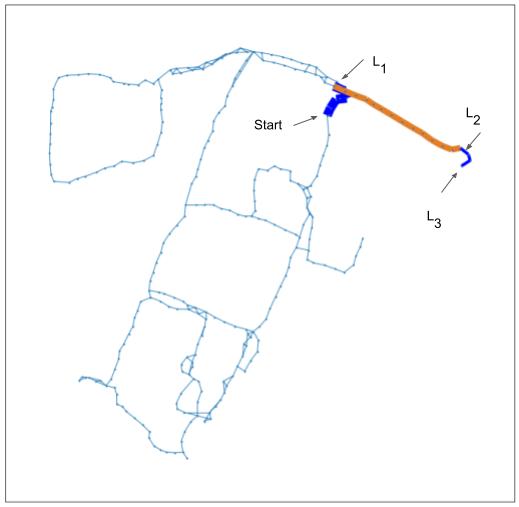}
    \end{subfigure}%
    \begin{subfigure}{.5\textwidth}
        \centering
        \includegraphics[width=0.9\textwidth]{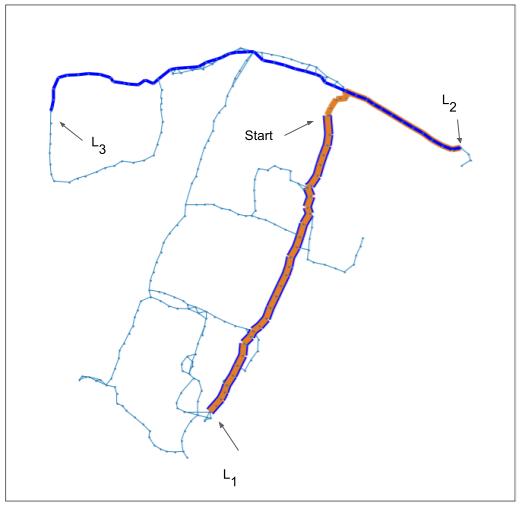}
    \end{subfigure}
    \caption{Examples of path planned by \sysName (left) and maximum likelihood planning (right). The start nodes and detected nodes are indicated with black arrows. In order to represent overlapping paths, we use colors interchangeably (start $\rightarrow L_1$: blue, $L_1 \rightarrow L_2$: orange, $L_2 \rightarrow L_3$: blue).  The path taken by \sysName is significantly shorter, resulting in a $5\times$ more efficient plan.}
    \label{fig:two_paths}
\end{figure}

\begin{figure}[t]
    \begin{subfigure}{.5\textwidth}
        \centering
        \includegraphics[width=0.9\textwidth]{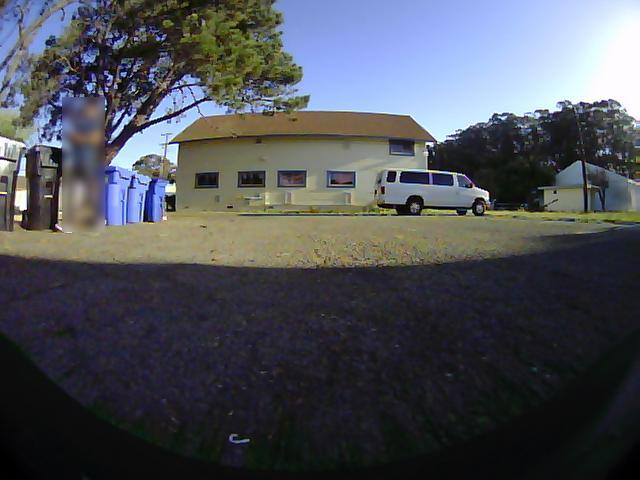}
    \end{subfigure}%
    \begin{subfigure}{.5\textwidth}
        \centering
        \includegraphics[width=0.9\textwidth]{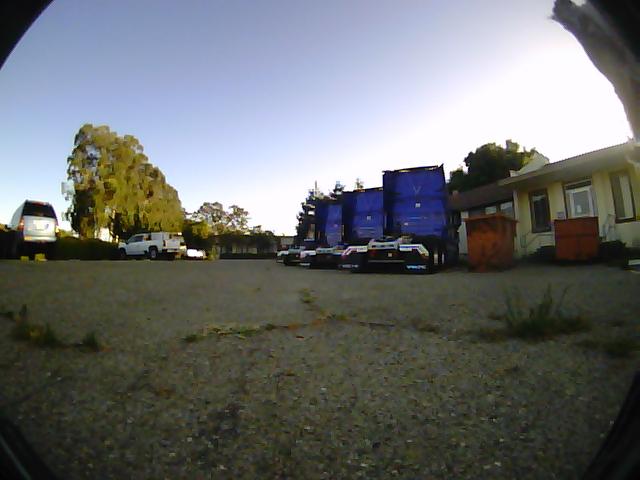}
    \end{subfigure}
    \caption{An example of failure to pick the correct image by maximum likelihood planning. Both images were selected for a prompt \textit{A photo of a blue dumpster}. The left one was selected as a part of the \sysName's graph search and the right was selected by maximum likelihood planning. In the latter case, the selected image contains a blue semi-truck and an orange trailer, but no blue dumpsters. This might be an example of an issue with the variable binding. The left image was edited to maintain anonymity.}
    \label{fig:failed_clip}
\end{figure}

Table~\ref{tab:max_likelihood} summarizes the performance metrics for the two planners. Unsurprisingly, the max likelihood planner suffers greatly in the form of efficiency, because it does not incentivize shorter paths (see Figure~\ref{fig:two_paths} for an example). Interestingly, the planning success suffers as well, especially in complex environments. Further analysis of these failure modes reveals cases where \vlm returns erroneous detections for some landmarks, likely due to the contrastive objective struggling with variable binding (see Figure~\ref{fig:failed_clip} for an example). While \sysName suffers from these failures as well, the second factor in the search objective $P_{t} (\bar{v})$ imposes a \emph{soft constraint} on the search space of the landmarks, eliminating most of these cases and resulting in a significantly higher planning success rate.

\subsection{Ablating the LLM}
\label{app:sec:llm_ablation}

As described in Section~\ref{sec:ablations} we run experiments comparing performance of different methods on extracting landmarks. Here we provide more details on the experiments. The source code to run this experiments is available in the file \texttt{ablation\_text\_to\_landmark.ipynb} in the repository (see Appendix~\ref{app:code_release}).

As the \textbf{metric of performance} we used average extraction success. For a query with a ground truth list of landmarks $L_{gt}$, where a method extracts list $L_{m}$, we define the methods extraction success as:
\[
    \frac{|\text{LCS}(L_{m}, L_{gt})|}{|L_{gt}|},
\]
where LCS is longest common subsequence and $|\cdot|$ denotes a length of a sequence or a list. This metric is measuring not only if correct landmarks were extracted, but also whether they are in the same order as in the ground truth sequence. When comparing landmarks we ignore articles, as we don't expect them to have impact on the downstream tasks.

All the experiments were run using APIs serving models. We used OpenAI's API  (\url{https://beta.openai.com/}) for GPT-3 and GooseAI (\url{https://goose.ai}) for the other open-source models. Both providers conveniently share the same API. We used the same, default parameters, apart from setting \texttt{temperature} to $0$: we don't expect that landmark extractions to require creativity and model's determinism improves reputability. For all the reported experiments, we used the same prompt as described in Appendix~\ref{app:prompt_engineering}. Please check out the released code for the exact prompts used.

\section{Code Release}
\label{app:code_release}
We released the code corresponding to the \llm interface, \vlm scoring, and graph search algorithm --- along with a user-friendly Colab notebook capable of running quantitative experiments from Section~\ref{sec:quantitative_analysis}. The links to the code and pickled graph objects can be found at our project page: \projectwebsite.

\section{Experiment Videos}
We are sharing experiment videos of \sysName deployed on a Clearpath Jackal mobile robotic platform --- please see \projectwebsite. The videos highlight the behavior learned by \sysName for the task of following free-form textual instructions and its ability to navigate complex environments and disambiguate between fine-grained commands.

\end{document}